\documentclass[letterpaper, 10 pt, conference]{IEEEtran}  

\IEEEoverridecommandlockouts                              

\usepackage{graphics} 
\usepackage{epsfig} 
\usepackage{mathptmx} 
\usepackage{times} 
\usepackage{amsmath} 
\usepackage{amssymb}  
\usepackage{kotex}
\usepackage{color}
\usepackage{multirow}
\usepackage{subfigure}

\usepackage{xcolor}
\usepackage[linesnumbered,ruled,vlined]{algorithm2e}
\SetKwComment{Comment}{/* }{ */}
\SetKwInput{KwData}{INPUT}
\SetKwInput{KwResult}{OUTPUT}

\usepackage[numbers]{natbib}
\bibpunct{[}{]}{,}{n}{}{;}
\usepackage[noadjust]{cite}  

\graphicspath{{./images/}}
\DeclareGraphicsExtensions{.pdf,.png,.jpg}

\DeclareMathOperator*{\argmin}{arg\,min}

\newenvironment{tightitemize}
{
	\begin{list}{$\bullet$}{%
			\setlength{\leftmargin}{8pt}
			\setlength{\topsep}{1pt}
			\setlength{\partopsep}{0pt}
			\setlength{\itemsep}{2pt}
			\setlength\labelwidth{5pt}}
		\ignorespaces}
	{\unskip\end{list}}
\newcounter{tecounter}
\newenvironment{tightenumerate}
{
	\begin{list}{\arabic{tecounter}\addtocounter{tecounter}{1}.}{%
			\setcounter{tecounter}{1}
			\setlength{\leftmargin}{12pt}
			\setlength{\topsep}{1pt}
			\setlength{\partopsep}{0pt}
			\setlength{\itemsep}{2pt}
			\setlength\labelwidth{7pt}}
		\ignorespaces}
	{\unskip\end{list}
}
\definecolor{orange}{rgb}{1,0.2,0}

\title{\LARGE \bf
Shape-adaptive Hysteresis Compensation for Tendon-driven Continuum Manipulators
}

\author{Young-Ho Kim and Tommaso Mansi  
	\thanks{Siemens Healthineers, Digital Technology \& Innovation,  Princeton, NJ, USA
		{\tt\scriptsize\{young-ho.kim,tommaso.mansi\}@siemens-healthineers.com}
	}%
}

\begin{document}
\maketitle
\begin{abstract}

Tendon-driven continuum manipulators (TDCM) are commonly used in minimally invasive surgical systems due to their long, thin, flexible structure that is compliant in narrow or tortuous environments. 
There exist many researches for precise tip control of the articulating section. However, these models do not account for the proximal shaft shape of TDCM, affecting the tip controls in practical settings.
In this paper, we propose a gradient-based shift detection method based on motor current that can easily find the offset of task space models ({\em i.e.,} hysteresis).
We analyze our proposed methods with multiple Intra-cardiac Echocariography catheters, which are typical commercial example of TDCM. Our results show that the errors from varied proximal shape are considerably reduced, and the accuracy of the tip manipulation is improved when changing external environmental structures.

\end{abstract}
	
\section{Introduction}

The tendon-driven continuum manipulator (TDCM) is widely used in many therapeutic\,\citep{daoud1999ep, khoshnam2017robotics, bai2012worldwide} and real-time diagnostic ({\em e.g.} endoscope\,\citep{ott11endoscopy,le16survey,dario03review}, colonoscope\,\citep{chen06kinematics,phee97locomotion}, and Intra-cardiac Echocardiography\,\citep{zhongyu21ice,kim2020automatic}) manipulators. 
The general physical shape of TDCM is a long, thin, flexible structure, which is consisted of four sections: a proximal handle, a proximal shaft, a bending section, and a distal tip. Multiple tendons are located in the rims of holes of the flexible backbone from the proximal handle to the distal tip. When pulling these tendons at the proximal handle, a load will be applied to the compliant backbone of the bending section and the corresponding segment will bend in the direction of the routed tendon.
These structural merits facilitate TDCMs to navigate in narrow and tortuous environments by adapting the shape of the proximal shaft, which makes them well-suited in minimally invasive treatment.


TDCM has many advantages and wide adoption for various application. 
Use of the TDCM in practical applications requires models of the bending shape in 3D space, which involves mapping the tendon lengths to the corresponding tip ({\em i.e.} backbone) position and orientation. However, TDCM is more complex than traditional rigid body robots. Many publications have presented a simplified approach with constant curvature assumption to obtain in closed-form forward kinematics\,\citep{xu08continuum,kim2020automatic,camarillo08tendon}. There already exist a handful of modeling reviews for tendon-driven continuum manipulators\,\citep{rao21tendon,webster10ccc}.
However, the varied proximal shaft shape was not accounted to the modeling of kinematics.


Moreover, the TDCM structure has a highly nonlinear behavior due to various slack, elasticity, and hysteresis phenomena in multiple coupled components involved in tendon control. To reflect dynamic characteristics, there exist several works\,\citep{do14hysteresis,xu17tendon,wang20hysteresis} related to friction/hysteresis compensation in tendon-driven manipulators. \citet{kato14tendon} proposed the forward kinematics mapping method to compensate tension propagation due to hysteresis operation.
Various mathematical models ({\em e.g.,} Bouc-Wen, Prandtl-Ishlinskii) have been proposed to model hysteresis in tendon-driven structures\,\citep{radu09identification,hassani13piezo,do14hysteresis}, which involved with many hyperparameters and associated with complicated identification procedures. Recently, \citet{lee21hysteresis} proposed a simplified hysteresis model for incorporating dead zone and backlash hysteresis phenomena, which are a commonly appeared in a commercial TDCM.

Although various attempts have been made for modeling, these models assume that the proximal shaft shape is a straight (or a fixed shape for one-time calibration), thus ignored the modeling errors in tip controls due to changed TDCM proximal shaft shape. These errors can be compensated by sensor feedback ({\em e.g.,} electromagnetic sensor, optical sensor, load cells). However, there are practical limitations (sterilization, cost, and size) which restrict adding traditional sensors to the tool tip to provide the necessary feedback for closed-loop control. To the best of our knowledge, our study is the first consideration of the shaft shape effect in task space models ({\em i.e.,} hysteresis).





\begin{figure}[t]
	\begin{center}
		\hspace{-5pt}
		\includegraphics[scale= 0.4]{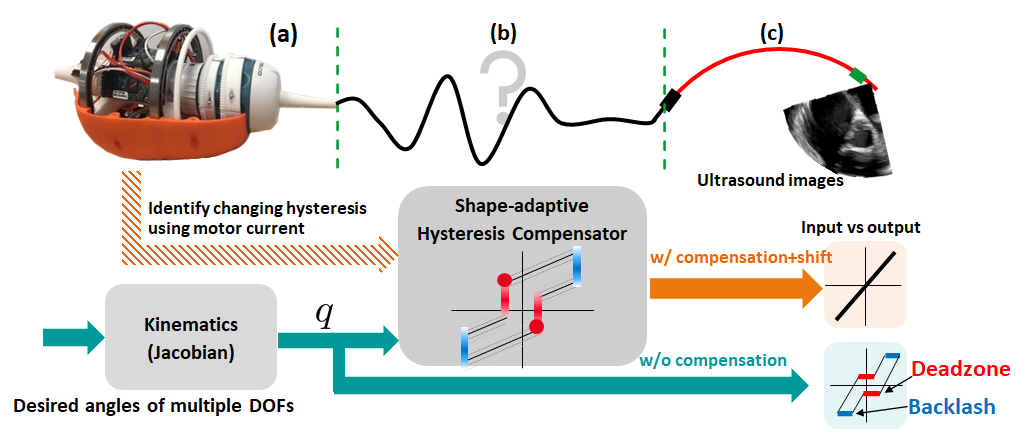}
		\caption{A representative diagram and scenario: Given desired pose at a given time, our goal is to identify changes of hysteresis in unknown shape of the proximal shaft, and compensate the configuration states, finding a compensated motion that minimizes errors.			
		The compensated configuration gives input equal to output while one without compensation generates dead zone and backlash. (a) [Proximal section] Intra-cardiac Echocardiography catheter is manipulated by robotic manipulators. The proximal section is a long, and thin flexible structure that can go through complicated vessel structures. (b) [Shaft shape] The catheter body structure can be located in narrow and tortuous environment during the procedure, so the flexible proximal shaft shape is arbitrarily changed, but mostly unknown. (c) [Distal section] This is the bending section. The precise tip control is most important problem that are widely investigated by many researchers. Our novel ideas are that 1) we use a motor current to detect and update shift of existing hysteresis without external sensors. 2) We analyze the effect of shaft shape, and evaluate our proposed method with varied inputs and various curvatures of the shaft. \label{fig:intro}}
		\vspace*{-20pt}
	\end{center}
\end{figure}

\begin{figure*}[t]
	\begin{center}
		\includegraphics[scale= 0.55]{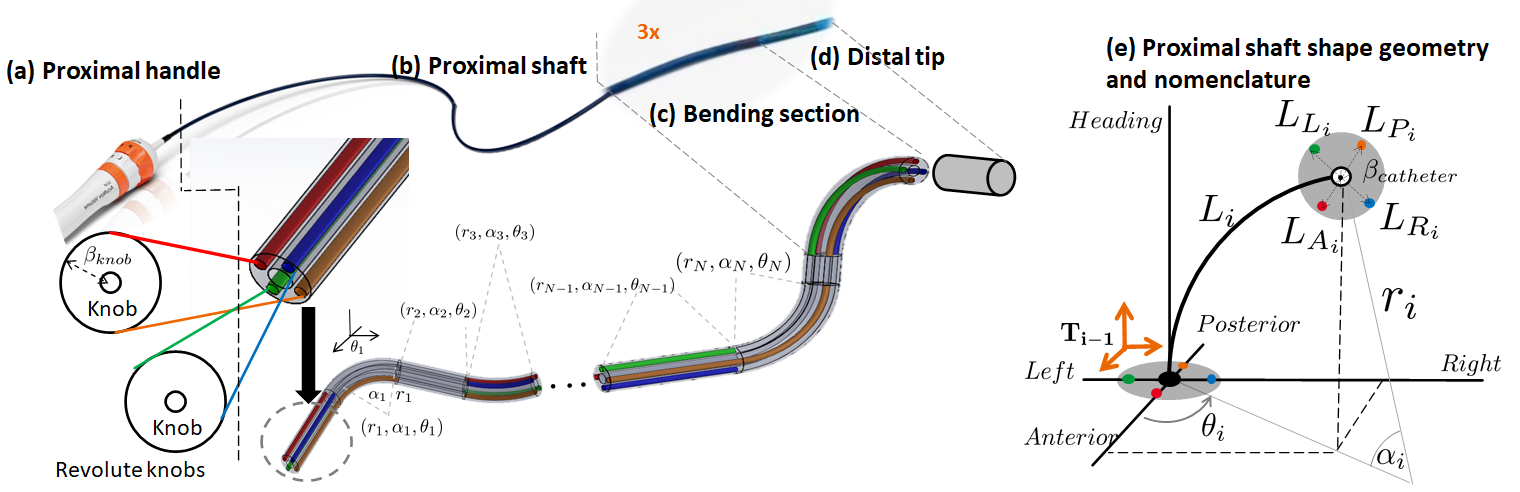}
		\caption{Illustrative examples of 2-DOF TDCM-based manipulator, Intra-cardiac Echocardiography catheter (ACUSON AcuNav, Source: Siemens Healthineers). There are four functional sections; (a) Two revolute knobs for each pair of threads (anterior(red)-posterior(orange) and right(blue)-left(green)) are in proximal handle section. (b) The proximal shaft is a long, flexible structure that can be compliant with anatomies. Thus, four threads can be pulled/released by environmental shape.  (c) This is a bending section for steering. (d) Ultrasound array is located at the distal tip. (e) The shape nomenclature of the proximal shaft
			\label{fig:structure}}
		\vspace*{-20pt}
	\end{center}
\end{figure*}
	
In this paper, we propose a simple methodology to identify shift of hysteresis measures due to the arbitrary shape change of the flexible proximal shaft.
More specifically,
(1) We propose a practical shape-adaptive hysteresis compensation based on the dead-zone detection using motor current.
(2) We introduce a systematic analysis how the TDCM proximal shape change affects the accuracy of tip controls.
(3) We evaluate our proposed method by periodic/non-periodic input motions for multiple tendon-driven catheters in varied shape of the catheter proximal shaft.

\section{Materials and Methods}

\subsection{Tendon-driven continuum manipulator}


The TDCM-based device is a long, thin, and flexible structure, which have one (or two) pair(s) of tendon-sheath pull mechanisms. The main structural functions can be decomposed into four sections, illustrated in Figure\,\ref{fig:structure}.  
\begin{tightenumerate}
	\item [{\bf Figure\,\ref{fig:structure}(a) [The proximal handle]}:]  Each pair is bound to a common knob, which can pull an individual thread by rotating the knob, allowing one thread to be pulled while the other remains passive. This structure has a highly nonlinear behavior due to various slack, elasticity, and hysteresis phenomena in multiple coupled components involved in tendon controls\,\citep{kim2020automatic,lee21hysteresis}. For example, the structure assumes ideally zero-slack transition between threads however, this is not realistically achievable in the proximal controls. 
	\item [{\bf Figure\,\ref{fig:structure}(b) [The proximal shaft]}:] There exist a larger diameter flexible backbone used for passage of tip sensors. Each tendon is located at the outer boundary of the backbone consist of a hollow polymer as a sheath and a thread sliding inside the sheath acting as a tendon. In Figure\,\ref{fig:structure}(b), four threads are located at 90~degree intervals; One DoF is anterior(red)-posterior(orange) threads, and another DoF is right(blue)-left(green) threads. This proximal shaft usually very long, thin, flexible structure ({\em e.g.,} Intra-cardiac echocadiography catheter, Transesophageal echocadiography catheter, etc.), thus the shape can be freely changed depending on anatomy shapes ({\em i.e.,} Air pathway, Inferior vena cava). 
	\item [{\bf Figure\,\ref{fig:structure}(c) [The bending section]}:]  This is an articulated section, controlled by pulling wires at the proximal handle. There exist various kinematics/jacobian models for bending section\,\citep{xu08continuum,kim2020automatic,webster10ccc} without consideration of the proximal shaft effect.
	\item [{\bf Figure\,\ref{fig:structure}(d) [The distal tip]}:] The distal tip section is used for various applications ({\em e.g.,} ultrasound, ablation, eletromagnetic pose tracking sensors).
\end{tightenumerate} 

We briefly review our motorized system that can manipulate multiple degree-of-freedom (DoF) tendon-driven continuum devices. The robot has four degrees of freedom; two DOFs for steering the tip in two planes (anterior-posterior knob angle $\phi_1$ and right-left knob angle $\phi_2$) using two knobs on the handle, and other two DOFs for bulk rotation and translation along the major axis of the catheter body. In this paper, we will focus on two knob controls depending on the TDCM shaft shape. We define the robot's configuration state, ${\bf q}=(\phi_1,\phi_2)$ in $\mathbb{R}^2$.


\subsection{The effect of the TDCM proximal shaft shape changes on tendon controls}

We define the segmented proximal shaft shape described in Figure\,\ref{fig:structure}(e). Let $L_i$ represent the $i$-th segmented backbone centerline length, where $i \in [1,N]$, $N$ is the total number of segmented part in the proximal shaft and $\sum_{i=1}^{N} L_i$ equal to the total length of the proximal shaft, ${ L_{shaft}}$. Let $L_{A_i}$, $L_{P_i}$, $L_{R_i}$ , and $L_{L_i}$ represent $i$-th segmented tendon length for anterior (red-colored), posterior (orange-colored), right (blue-colored), and left (green-colored), respectively. $\alpha_i$ is the angle of the curvature for $i$-th segment. $\theta_i$ is the right-handed rotation angle from the anterior-axis of $T_{i-1}$ along the heading-axis of $T_{i-1}$. $r_i$ represents the radius of the curvature of the backbone. Then, we define $i$-th segment of the proximal shaft $S_i= (r_i, \alpha_i, \theta_i)$. Additionally, Let $\beta_{catheter}$ be the distance between the center backbone and each thread, and $\beta_{knob}$ represents the revolute knob radius.

Then, $L_{A_i}$, $L_{P_i}$, $L_{R_i}$, and $L_{L_i}$ can be computed as follows;
\begin{equation}\label{eq:shape_segment}
{\small
\begin{aligned}
L_{A_i} = L_i - \alpha_i \beta_{catheter} \cos(\theta_i),\\
L_{P_i} = L_i + \alpha_i \beta_{catheter} \cos(\theta_i),\\
L_{R_i} = L_i - \alpha_i \beta_{catheter} \sin(\theta_i),\\
L_{L_i} = L_i + \alpha_i \beta_{catheter} \sin(\theta_i).
\end{aligned}
}
\end{equation}

Then, the overall length of each tendon before the bending section will be as follows;
\begin{eqnarray}\label{eq:shape_total}
{\small
{\bf L}_A = \sum_{i=1}^{N}L_{A_i},~
{\bf L}_P = \sum_{i=1}^{N}L_{P_i},~
{\bf L}_R = \sum_{i=1}^{N}L_{R_i},~
{\bf L}_L = \sum_{i=1}^{N}L_{L_i}.
}
\end{eqnarray}

The total amount of the pulled length for each thread before the bending section is defined as $(\Delta{\bf L}_A, \Delta{\bf L}_P, \Delta{\bf L}_R, \Delta{\bf L}_L)$ computed as follows;
\begin{equation}\label{eq:shape_deflection}
{\small
\begin{aligned}
\Delta{\bf L}_A =  \sum_{i=1}^{N} \alpha_i \beta_{catheter}  \cos(\theta_i),~
\Delta{\bf L}_P = -\sum_{i=1}^{N} \alpha_i \beta_{catheter}  \cos(\theta_i),~\\
\Delta{\bf L}_R =  \sum_{i=1}^{N} \alpha_i \beta_{catheter}  \sin(\theta_i),~
\Delta{\bf L}_L = -\sum_{i=1}^{N} \alpha_i \beta_{catheter}  \sin(\theta_i).
\end{aligned}
}
\end{equation}

Finally, the total pulled length of each thread at each knob ($\phi_1$, $\phi_2$) from the proximal handle to the bending section is defined as follows; 
\begin{equation}\label{eq:shaft_bending}
\begin{aligned}
\phi_1 \beta_{knob} = \left\{\begin{array}{ll}
\Delta{\bf L}_A + \Delta L_{A_{bending}} &{ \textrm{if}~ \phi_1>0,} \\
\Delta{\bf L}_P + \Delta L_{P_{bending}} &{ \textrm{otherwise}~ }
\end{array} \right. \\
\phi_2 \beta_{knob} = \left\{\begin{array}{ll}
\Delta{\bf L}_R + \Delta L_{R_{bending}} &{ \textrm{if}~ \phi_2>0,} \\
\Delta{\bf L}_L + \Delta L_{L_{bending}} &{ \textrm{otherwise}~ }
\end{array} \right.
\end{aligned}
\end{equation}
where $\Delta L_{A_{bending}}$, $\Delta L_{P_{bending}}$, $\Delta L_{R_{bending}}$, $\Delta L_{L_{bending}}$ represent pulled thread lengths in the bending section, which are used for the input of forward/inverse kinematics. 

Herein, we are interested in how to realize $\Delta{\bf L}$ for each thread, which affects the total pulled length of threads. Therefore, we do not re-visit fundamental forward/inverse kinematics in this paper. The detailed kinematics models can be found in \,\citep{webster10ccc,rao21tendon}. 
Most literature underestimate the effect of the proximal shape changes due to varied environmental structures. In practice, $\Delta{\bf L}_A$, $\Delta{\bf L}_P$, $\Delta{\bf L}_R$, and $\Delta{\bf L}_L$ might not be zero. Then, there exist additional tensions associated with the amount of pulled length due to the proximal shaft shape. This affects the initial state of the bending section ${\bf q}_{init}$, which might not be $(0, 0)$. So, we need to find non-zero ${\bf q}_{init}$ and compensate changes for a precise control of the tip.

Then, our problem is that given desired pose at a given time, our goal is to identify changes of hysteresis in unknown shape of the proximal shaft, and compensate the configuration states ${\bf q}'$, finding a compensated motion that minimizes errors {\bf J}. Then, the problem follows:
\begin{equation}\label{eq:problem}
\begin{aligned}
{\bf q}' = \argmin_{{\bf q}_{init} \in \mathbb{R}^n} {\bf J}({\bf q},{\bf q}_{init})
\end{aligned}
\end{equation}


Since we focus on modeling from the proximal handle to the bending section, we need to consider frictional behaviors caused by: 1) backlash hysteresis due to friction forces between the sheath and tendons, 2) dead zone due to structural wire slack in the driving parts. These factors contribute to the degradation of control accuracy and limit the potential performance of robotic controllers for the off-the-shelf TDCM-based devices.

Moreover, Equation\,\eqref{eq:shaft_bending} shows that the existed curvature shapes affect the total deflection of the bending section. To measure $S_i$ for Equation\,\eqref{eq:shape_deflection}, external sensors might be required such as Fiber Bragg Grating for a direct shape sensing or images (optical, CT/MRI) for detection/segmentation.
However, there exist a practical limitation due to cost and size of sensors. Thus, our focus for the control strategy is open loop with no spatial feedback for the shape. Instead, we investigate how to use motor current for identifying changes of hysteresis from the proximal handle side.

\subsection{Shift controls of hysteresis compensation}

Before we explain more detailed of our proposed algorithm, we briefly explain our previous work\,\citep{lee21hysteresis} (a simplified hysteresis model) because that modeled dead zone and backlash hysteresis phenomena together, and our proposed method associated with dead zone detection. However, our method can be applicable for any hysteresis model that handles dead zone in a straight condition of the shaft.

\subsubsection{Review of a simplified hysteresis model}

\begin{figure}[t]
	\begin{center}
		\includegraphics[scale= 0.5]{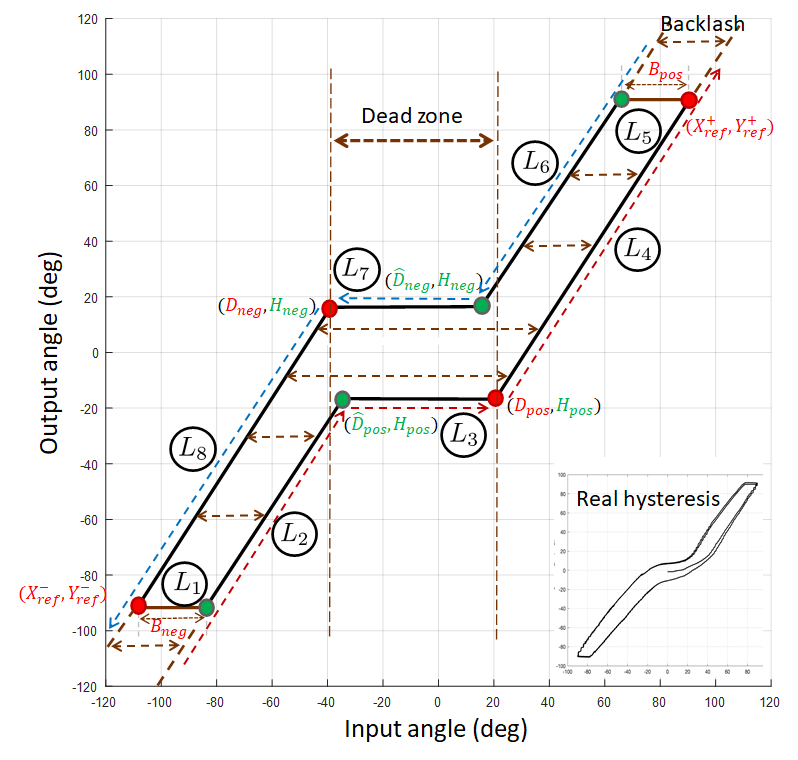}
		\caption{ This is an overall diagram to explain one existing hysteresis model\,\citep{lee21hysteresis} that can treat both dead zone and backlash together. There exist four main parameters in red colored: the size of the backlash hysteresis ($B_{pos}$, $B_{neg}$), and the range of the dead zone ($D_{pos}$, $D_{neg}$), which can be acquired from the motor current. Other associated parameters, $\hat{D}_{pos}$, $\hat{D}_{neg}$, $H_{pos}$, and $H_{neg}$ can be computed based on known main parameters. 
			\label{fig:hysteresis}}
	\end{center}
\end{figure}

In our previous work\,\citep{lee21hysteresis}, we proposed a piecewise linear approximation to represent the non-linear hysteresis phenomenon ({\em i.e.,} both dead zone and backlash). 
For completeness, we summarize the hysteresis model.
This model consisted of a total of eight linear equations, half of which are when the velocity is positive and others are when the velocity is negative.

{\small
	\begin{eqnarray}\label{eq:D_hats}
	\begin{aligned}
	\hat{D}_{pos} &= (H_{pos}-H_{neg})/\omega + D_{neg}+B_{neg} \\
	\hat{D}_{neg} &= (H_{neg}-H_{pos})/\omega + D_{pos}-B_{pos}
	\end{aligned}
	\end{eqnarray}
}

{\small
	\begin{equation}\label{eq:model_eqs}
	L(x_t,\dot{x}_t,\dot{x}_{t-1}) =
	\begin{cases}
	\text{$L_{1}$}:     \omega(-X_{ref}-D_{neg})+H_{neg} \\
	\text{$L_{2}$}: \omega(x-\hat{D}_{pos})+H_{pos}\\
	\text{$L_{3}$}: H_{pos}\\
	\text{$L_{4}$}: \omega(x-D_{pos})+H_{pos}\\
	\text{$L_{5}$}: \omega(X_{ref}-D_{pos})+H_{pos}\\
	\text{$L_{6}$}: \omega(x-\hat{D}_{neg})+H_{neg}\\
	\text{$L_{7}$}: H_{neg}\\
	\text{$L_{8}$}: \omega(x-D_{neg})+H_{neg}\\
	\end{cases}
	\end{equation}
}

To define a finite collection of linear functions, we define $x_t\in {\bf q}$ and $\dot{x}_t\in {\bf \dot{q}}$ as the one-DOF input state and velocity at time $t$, respectively, where $t$ is a temporal index. 
There are four main parameters; $D_{pos}$ and $D_{neg}$ represent the dead zone range for positive and negative directions. $B_{pos}$ and $B_{neg}$ represent the size of the backlash hysteresis for each direction. 
Six associated parameters are denoted; $H_{pos}$ and $H_{neg}$ represent the height of the dead zone for each direction. $\hat{D}_{pos}$ and $\hat{D}_{neg}$ represent the opposite side of the dead zone $D_{pos}$ and $D_{neg}$, respectively. Additionally, $\omega$ denotes the slope of the lines, and let $(X^+_{ref}, Y^+_{ref})$ and $(X^-_{ref}, Y^-_{ref})$ denote reference points for each direction, which is the input and the real output state. 
Figure\,\ref{fig:hysteresis} shows how eight equations are divided and switched over the region with all parameters.

Then, This hysteresis model $L(x_t,\dot{x_t},\dot{x}_{t-1})$ (Equation\,\eqref{eq:model_eqs}) is determined by eight linear equations with four known parameters from motor current at the straight condition, while other parameters are also derived from known parameters in Equation\,\eqref{eq:D_hats}. This model does not consider the shape change of the proximal shaft. More detailed dynamic transition rules among piece-wise models are described in \,\citep{lee21hysteresis}.

\subsubsection{Experimental validation for dead zone and motor current} \label{sec:deadzone_motorcurrent}


We conduct a systematic test where the desired input is a simple sweep motion in the form of a sine wave that has been commonly used in other studied\citep{hassani13piezo, do14hysteresis,xu17tendon}.
The sweeping motion angle range is ${\pm 40^{\circ}}$. Two cycles of sweeping motions are applied with $40^{\circ}/sec$, and we design three cases of the shape of the sheath; (1) the straight condition $\alpha = 0$, (2) Right-side bent ($\theta = 90^{\circ}$, $\alpha = 90^{\circ}$, $r = 100~mm$) (3) Left-side bent ($\theta = -90^{\circ}$, $\alpha = 90^{\circ}$, $r = 100~mm$). Then, we collect the following data for 1-DoF: (1)~the desired robot configuration input, $\phi_1$ (or $\phi_2$), (2)~the real output angle of the bending section using electromagnetic tracker with inverse kinematics, and (3)~the motor current $C$ acquired from motor drivers in real time. We applied the proper filter for all measures (3rd order Butterworth filter, cutoff frequency 20 hz). 

\begin{figure}[t]
	\begin{center}
		\includegraphics[scale= 0.45]{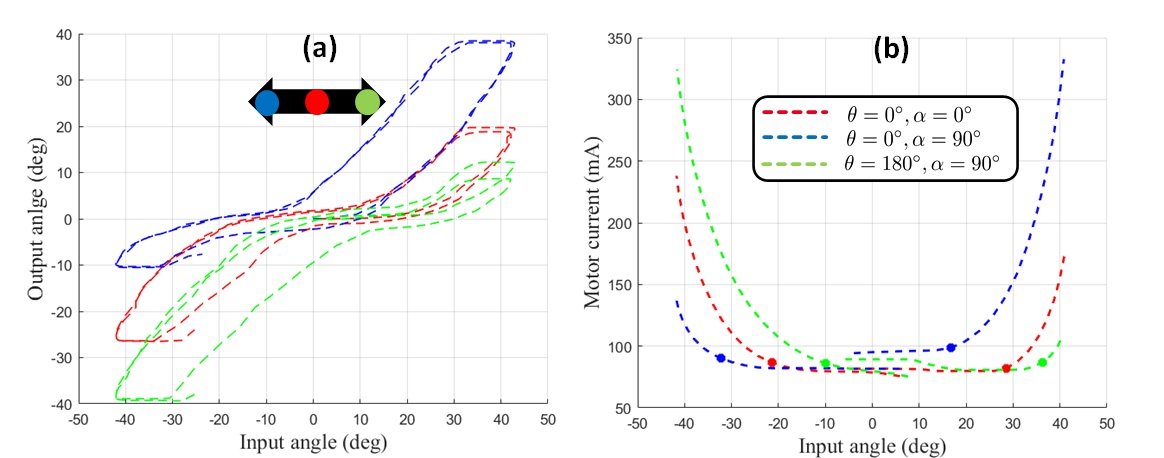}
		\caption{ One analytical data shows how the same input shows different results at the different shape conditions. The input is $\pm 40^\circ$ sinusoidal function. 
		The red-dotted line is a straight condition ($\theta=0$, $\alpha=0$), the blue-dotted line is a $90^{\circ}$ bent ($\theta=0$, $\alpha=90$) in anterior direction, and the green-dotted line is a $90^{\circ}$ bent ($\theta=180$, $\alpha=90$) in posterior direction. (a) the desired robot state versus the real output angle measured by EM tracking system (b) the desired robot state versus the motor current. The big dots indicate the estimated dead zone values ($\tilde{D}_{pos}$, $\tilde{D}_{neg}$) by our proposed method. Overall, the shift of hysteresis happened positive or negative direction depending on bending direction.
			\label{fig:relationship}}
	\end{center}
\end{figure}

Figure\,\ref{fig:relationship} shows the result of the systematic test to understand how the shape of the shaft affects the output of the tip control at the existing hysteresis model.
Figure\,\ref{fig:relationship}(a) shows the desired robot state versus the real output angle measured by EM tracking system.
Figure\,\ref{fig:relationship}(b) shows the desired robot state versus the motor current. Overall, the basic hysteresis (red-dotted line, a straight condition) of Figure\,\ref{fig:relationship}(a) is shifted to the positive (green-dotted line, Right-side bent) or negative direction (blue-dotted line, Left-side bent) depending on bending direction.

\subsubsection{Algorithm for shape-adaptive hysteresis compensation}

When we change the shape condition in the proximal shaft, the hysteresis phenomena is shifted both X-axis direction. Then, all linear equations of models will be shifted by ${\bf q}_{init}$.

Basically, we assume that the dead zone $D$ might be a key point to detect a shift of hysteresis by the shape changes because it is due to structural wire slack in the proximal handle; One thread will be pulled, while opposite side of the thread will be released, thus the knob will be balanced with a certain level of shift.
However, the backlash $B$ phenomena comes due to the friction between sheath and thread, which might be independent with the shift phenomena due to the shape of the shaft structure.

We also assume that only dramatically changing point of motor current happened when the wire pulled in non-dead zone, which is shown in Figure\,\ref{fig:relationship}(b). For example, even if the knob is manipulated in dead zone, the real measure of pose will not be changed because tension is not applied to the thread, thus the change of motor current will be very small. This mechanical properties give an intuition to solve our problem. Our motor current is continuous and differentiable function, and does not have a local minima. This is a true if there is no interaction with environment when we find the key point of dead zone by motor current.
Thus, there is only a unique solution for each side of dead zone. We specify the range of the unique solution. 

To estimate the newly updated dead zones ($\tilde{D}_{pos}$, $\tilde{D}_{neg}$) for each knob, we introduce a simple gradient-based method in Algorithm\ref{alg:gradient}. Let $\epsilon_{upper}, \epsilon_{lower}$ be the upper/lower bound of the gradient, which will determine aggressive or conservative level of search. $\bf{u} \in U$ be the unit step of control. In the search phase, the unit motion will be applied while the derivative of the motor current $\nabla C$ is computed. Then, based on $\epsilon$ bound, we can determine the change of point for both side. If $\nabla C$ is greater than $\epsilon_{upper}$, then the motion $direction$ will be changed. If $\nabla C$ is greater than $\epsilon_{lower}$, then $offset$ will be computed depending on $direction$. The $offset$ will be applied to both $\tilde{D}_{pos}$ and $\tilde{D}_{neg}$. In our approach, we do not need to sweep both side of dead zone, we simply move to one direction and find the closest offset. One real example is shown in Figure\,\ref{fig:relationship}(b) that applied our proposed method and estimated new dead zone values ($\tilde{D}_{pos}$, $\tilde{D}_{neg}$) for three cases of sweeping; The big dots (red, green, blue) are indicating the new dead zone due to shape changes.

\begin{algorithm}[h]
	\caption{Gradient$\_$based$\_$Shift$\_$Detection}\label{alg:gradient}
	\KwData{${D}_{pos}, {D}_{neg},  \epsilon_{upper}, \epsilon_{lower}$}
	\KwResult{$\tilde{D}_{pos}$, $\tilde{D}_{neg}$}
	$offset = 0$, $direction = 1$ \tcp*[r]{Initialization}
	\While{MAX-Iteration \& $offset = 0$}{
		\tcp{First: apply a step motion \bf{u}}
		${\bf q}(t) = {\bf q}(t-1) + direction\cdot \bf{u}$\;
		\tcp{Second: compute the slope of motor current C}
		$\nabla C = \frac{dC}{dq}$\;
		$offset = 0$\;
		\uIf{if $\nabla C >= \epsilon_{upper}$}{
			\tcp{Change direction of $u$}
			direction -=direction\;
		}
		\uElseIf{$\nabla C>= \epsilon_{lower}$}{
			\tcp{Update $\tilde{D}$}
			\uIf{direction $\geq 0$}{				
				$offset = {\bf q}(t) - {D}_{pos}$ \;
				$\tilde{D}_{pos} = {D}_{pos} + offset$\;
				$\tilde{D}_{neg} = {D}_{neg} + offset$\;				
			}
			\Else{
				$offset = {\bf q}(t) - {D}_{neg}$ \;
				$\tilde{D}_{pos} = {D}_{pos} + offset$\;
				$\tilde{D}_{neg} = {D}_{neg} + offset$\;
			}
			break\;
		}
	}
\end{algorithm}

Our algorithm is a real-time method, but requires the pure motor current to detect at least one side of dead zone with associated behavior motions. Thus, our method can not be applied with other purpose of motions. Also, we can not detect a dynamic dead zone in changing environment in real-time. 

Since this method is not automatically detecting shape changes during the procedures, there are several cases that we can activate our algorithm in semi-automation (or manually). (1) Most of cases are mainly controlling two knob ($\phi_1$ and $\phi_2$) when the TDCM body arrived at the target area. If new motions are only associated with two knob controls. Then, we can simply update the current shift of hysteresis by our method. (2) If there exist only bulk translation of a long TDCM structure, then there possibly exist new segments of curvature, which requires to apply our method in order to update shift of hysteresis. (3) If there exist a bulk rotational motion, then there is no new segments of curvature assuming environments are fixed and there is no twist effect on the proximal shaft. In this case, we can simply apply our method to update parameter. Otherwise, the existing shift values could be re-calculated based on Equation\,\ref{eq:shape_deflection} without additional motions to update the shift because we know how much the TDCM device rotated ($\theta$ in global coordinate system) from motor state. Thus, it can be simply adjusted by the amount of the body rotation. This is not covered in this paper.

\section{Experiment and Result}

\begin{figure*}[t]
	\subfigcapskip = -8pt
	\begin{center}		
		\subfigure[Time  versus  output  angle (1-DoF) at a straight condition ($\alpha = 0^\circ$)]{\includegraphics[scale=0.33]{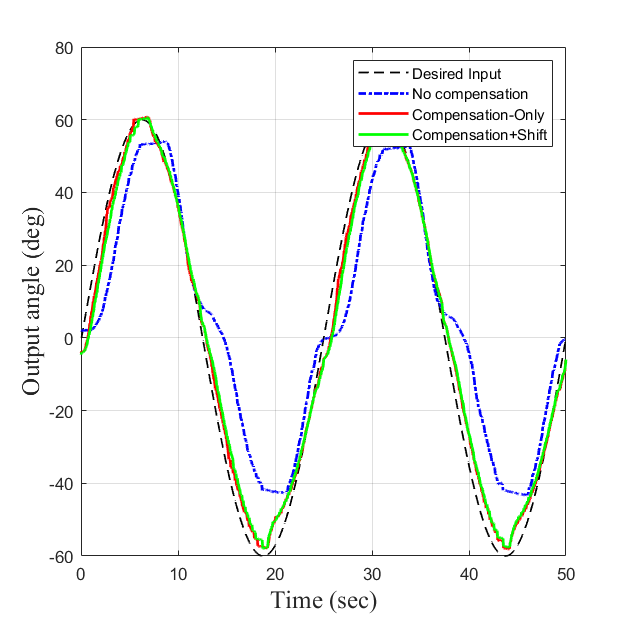}\label{fig:1AP}}
		\subfigure[Time  versus  output  angle (1-DoF)  at ($\alpha = 45^\circ$)]{\includegraphics[scale=0.33]{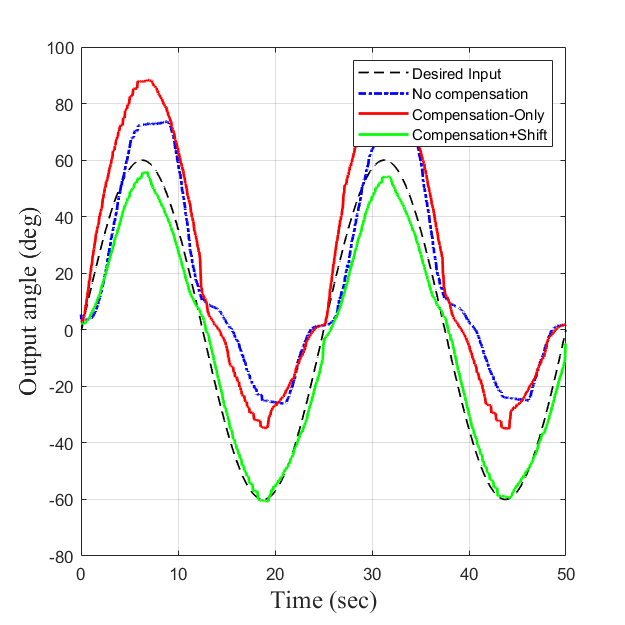}\label{fig:1LR}}
		\subfigure[Time  versus  output  angle (1-DoF) at ($\alpha = 90^\circ$)]
		{\includegraphics[scale=0.33]{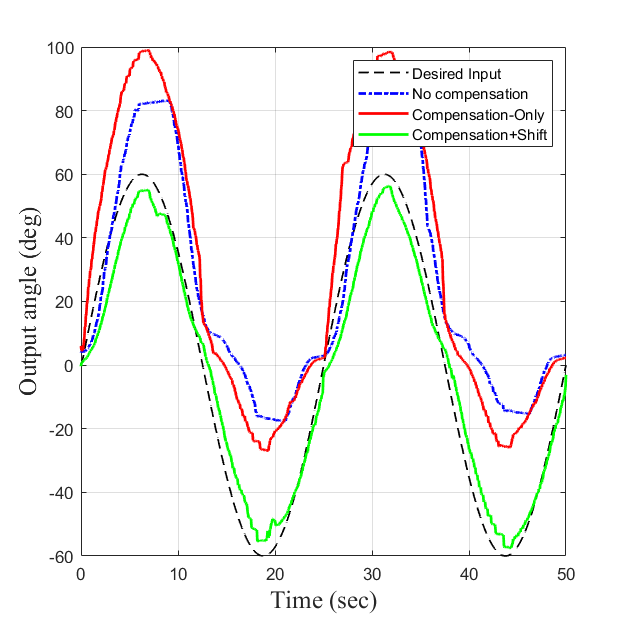}}\\		
		\subfigure[Input  versus  output angle,  $\phi_1$]{\includegraphics[scale=0.33]{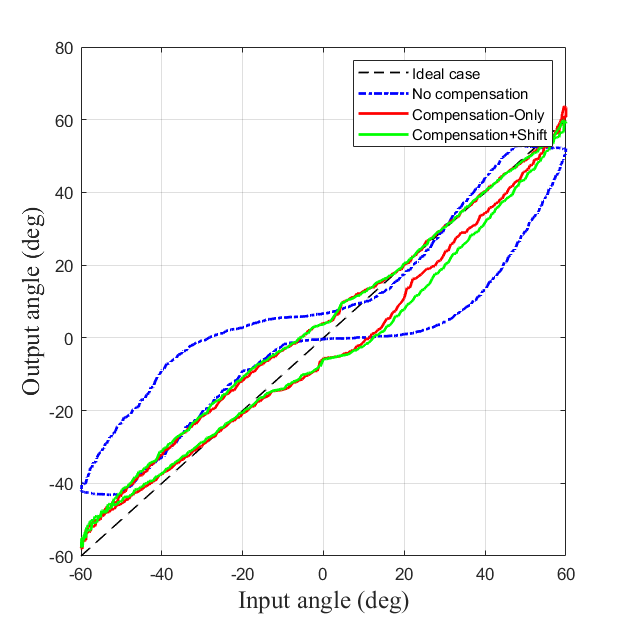}}
		\subfigure[Input  versus  output angle,  $\phi_1$]{\includegraphics[scale=0.33]{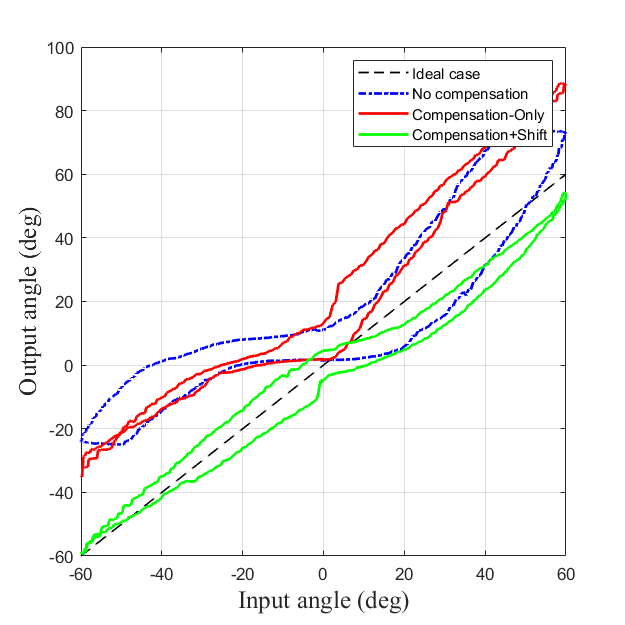}}
		\subfigure[ Input  versus  output angle, $\phi_1$]{\includegraphics[scale=0.33]{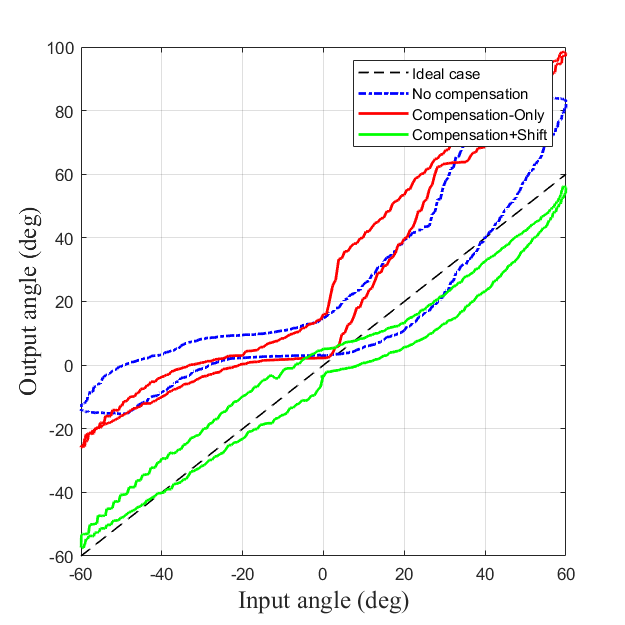}}
		\caption{ This shows 1-DoF results of one catheter for three shape scenarios ($\alpha$ = [$0$, $45$, $90$]) with a periodic input: The black dot is the ground truth). The blue dot is {\it No Compensation}. The red is {\it Compensation-Only}. The green is our proposed method {\it Compensation+Shift}. The first column is a straight condition. The second column is $\alpha = 45^\circ$. The third column is $\alpha = 90^\circ$. The first row shows time versus output angle ($\phi_1$ or $\phi_2$). The second row shows the input angle versus the output angle for $\phi_1$ (or $\phi_2$) motions.
			\label{fig:result_1dof_total}}
	\end{center}
\end{figure*}

We used ICE catheter as an example of TDCM attached to motorized system (shown in Figure\,\ref{fig:intro}) that can manipulate two knobs $\phi_1$ and $\phi_2$.
To validate our proposed method, we used two electromagnetic sensors (Model 800 sensor, 3D guidance, Northern Digital Inc.) attached to the both ends of bending section. Thus, we get real-time position and orientation of the tip as a ground truth via inverse kinematics. 
 
First, we carried on parameter identification for one hysteresis model for each catheter at the straight shaft condition. We used motor current to detect $D_{pos}$, $D_{neg}$, $B_{pos}$, and $B_{neg}$. We used $\omega$ as 1.45 as the average value, then all other parameters can be identified (More detailed in \citep{lee21hysteresis}).
Then, we have one hysteresis model for each catheter at the straight shaft condition.

\subsection{Experimental Setup: controllers, shape scenarios, input motions, conditions}
Two ICE catheters were tested as following scenarios with three trials.

\begin{tightitemize}
	\item[\bf Controllers:] Three types of controllers are considered for comparison of the experimental results; (1) {\it No compensation:}~This controller is designed to send ${\bf q}$ directly to the motors without any compensation. (2) {\it Compensation-Only:}~This is applying the original hysteresis compensation method, no matter the shaft shape changed. (3) {\it Compensation+Shift:}~This method applied our proposed method to the original hysteresis method.
	\item[\bf Shape Scenarios:] Three shape scenarios are tested; (1) straight condition (2) $\alpha = 45^{\circ}$ (3) $\alpha = 90^{\circ}$. We used a specialized holding structure that can easily set $\alpha$ for both $45^{\circ}$ and $90^{\circ}$ with a stable holder for the shaft.
	\item[\bf Two types of motions] Two types of input motions are designed to validate our proposed methods. (1) Periodic: A
	periodic motion is designed as $60$ amplitude and $0.04$~Hz.
	(2) Non-periodic: A non-periodic motion was designed as
	a combination of two sinusoidal signals with $30$ for both
	amplitudes and $0.02$~Hz and $0.02\sqrt{3}$~Hz in frequencies.
	\item[\bf Operating conditions:] Two operating conditions are designed. (1) One-DoF operation: We used both AP bending section and LR bending section. (2) Two-DOFs operation;  We apply the same inputs simultaneously for both knobs, but one knob is twice faster than another knob. We only tested two-DOFs for non-periodic inputs.
\end{tightitemize}

 We applied our method for each shape scenario to identify shift of dead zone. Then, we applied our updated dead zone, and tested with two types of motions and different operating conditions assuming the curvature of the shaft shape associated with our scenario is fixed.

 To evaluate our proposed method, we use the magnitude of the peak-to-peak error (PTPE), which is measured between the highest value and the lowest errors. In addition, we use the root mean squared error (RMSE) to show errors. Both PTPE and RMSE are shown with mean and standard deviation ($\mu$, $\sigma$) for repeatability. The improvement rate shows how much errors can be reduced compared to {\it No Compensation} and {\it Compensation Only}.

\subsection{Result of the non-linear hysteresis compensation}

	\begin{table*}[t]
	\centering
	\caption{PTPE and RMSE for 1-DoF: periodic  inputs}\label{table:1d_all}
	\scalebox{1.2}{\begin{tabular}{|c||c|c||c|c||c|c|}
			\hline                {\bf Test inputs} & \multicolumn{6}{|c|}{\bf Periodic 1D-Inputs} \\
			\hline				{\bf Shaft shape}			&\multicolumn{2}{|c||}{Straight ($\alpha = 0^\circ$)}
			&\multicolumn{2}{|c||}{Small curvature ($\alpha = 45^\circ$)}
			&\multicolumn{2}{|c||}{Large curvature ($\alpha = 90^\circ$)} \\
			\hline
			{\bf Metrics}             & {\bf PTPE}    & {\bf RMSE}  & {\bf PTPE}     & {\bf RMSE}  & {\bf PTPE}     & {\bf RMSE} 
			\\ \hline
			No Compensation (deg)   
			&    (57.2, 4.2)             &  (18.5, 3.4)       &    (61.9, 2.1)             &  (25.4, 1.5) 
			&    (75.0, 10.1)             &  (38.2, 5.5)      \\ \hline
			Compensation-Only (deg) 
			&   (18.6, 2.3)             &   (7.7, 1.5)     
			&    (30.12, 2.5)             &   (24.2, 3.5)  
			&    (45.0, 5.6)             &   (31.3, 4.1)    \\ \hline
			Compensation+Shift (deg) 
			&   -          &   -    
			&    (24.3, 2.3)             &   (7.9, 1.3)     
			&    (26.0, 3.2)             &   (8.0, 1.2)     \\ \hline
			Improvement rate ($\%$) &   -             &   -   &   60.74 and 19.3             &   68.9 and 67.3   
			& 65.3 and 42.2   &   79 and 74.4                \\    
			\hline
	\end{tabular}}
\end{table*}

\begin{table}
	\centering
	\caption{RMSE for 2-DoF: non-periodic inputs}\label{table:2d_all}
	\scalebox{0.85}{\begin{tabular}{|c||c|c|}
			\hline        {\bf Test inputs} & \multicolumn{2}{|c|}{{\bf Non-Periodic 2D-Inputs}} \\
			\hline
			{\bf Shaft shape}      &  \multicolumn{2}{|c|}{Two curvatures($\alpha_1 = 45^{\circ}$, $\theta_1 = 0$, $\alpha_2 = 45^{\circ}$, $\theta_2 = 90$)}  	\\ \hline
			{\bf Metrics}             & {\bf PTPE}    & {\bf RMSE}   	\\ \hline
			{ No Compensation (deg)}   
			&    (55.6, 8.3)  &    (18, 3.2)                   \\ \hline
			{ Compensation-Only (deg)} 
			&   (30.5, 3.9)  &   (16.0, 3.3)                    \\ \hline
			{ Compensation+Shift (deg)} 
			&   (25.1, 3.4)  &   (10.1, 3.7)                     \\ \hline
			Improvement rate ($\%$) 
			&   45.8 and 17.7        &   44 and 37.5                    \\   
			\hline
	\end{tabular}}
\end{table}

We show the overall performance evaluation for 1-DoF with a periodic input in Table\,\ref{table:1d_all}, which shows PTPE and RMSE for each shape scenarios and each controller. As the angle of shaft shape is increased, both {\it No Compensation} and {\it Compensation-Only} errors are significantly increased, while {\it Compensation+Shift} keeps improving regardless of any shape condition. Our proposed method improved PTPE by $60\%$ to $65\%$ and RMSE $69\%$ and $79\%$ for {\it No Compensation}, while improved PTPE by $20\%$ to $42\%$ and RMSE $67\%$ and $74\%$ for {\it No Compensation}. This informs that any compensation without consideration of the shaft shape factor might be worsen the performance. 

Figure\,\ref{fig:result_1dof_total} shows 1-DoF results of one catheter.
The first row shows time versus 1-DoF output angle (either $\phi_1$ or $\phi_2$). The second row shows the input angle versus the output angle for (either $\phi_1$ or $\phi_2$).
The first column is tested at the straight shaft shape condition ($\alpha = 0$). The second column is at the small curvature $\alpha = 45^\circ$, and the last column is at the large curvature $\alpha = 90^\circ$. From left to right, we can see that errors of {\it No compensation} and {\it Compensation Only} are significantly increased. However, {\it Compensation+Shift} shows a great improvement.

\begin{figure}[t]
\subfigcapskip = -8pt
	\begin{center}		
		\subfigure[Time  versus  output  angle, $\phi_1$]
		{\includegraphics[scale=0.25]{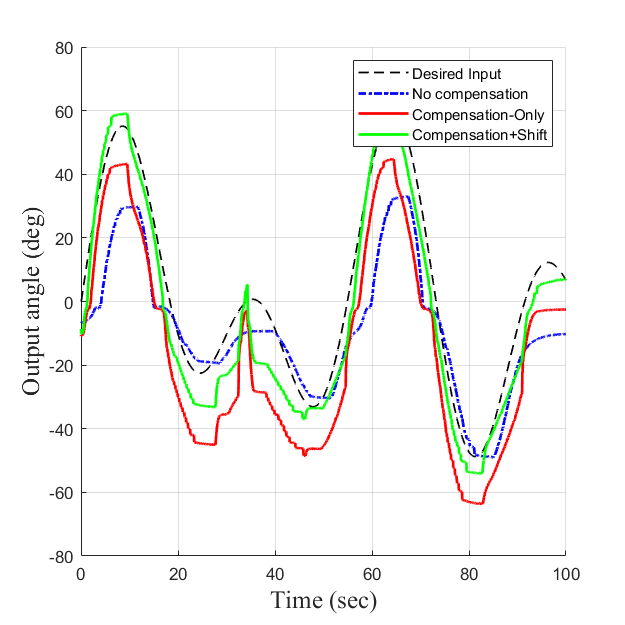}}
		\subfigure[Time  versus  output  angle, $\phi_2$]
		{\includegraphics[scale=0.25]{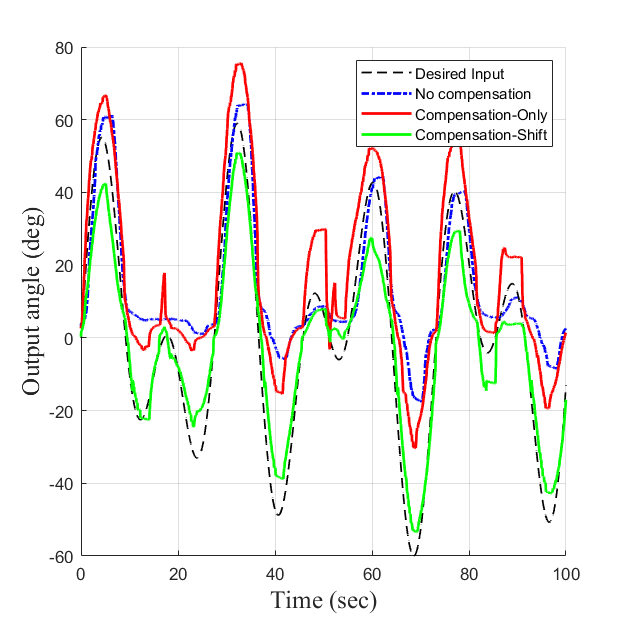}}		
		
		\subfigure[ Input  versus  output angle, $\phi_1$]{\includegraphics[scale=0.25]{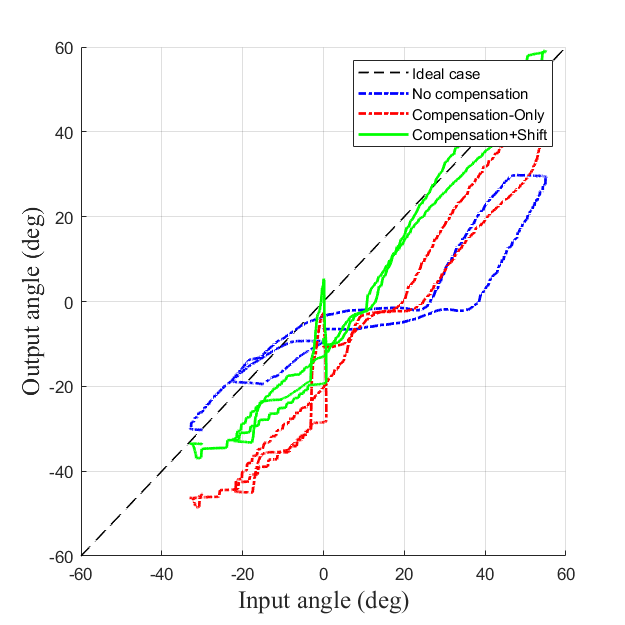}}
		\subfigure[ Input  versus  output angle, $\phi_2$]{\includegraphics[scale=0.25]{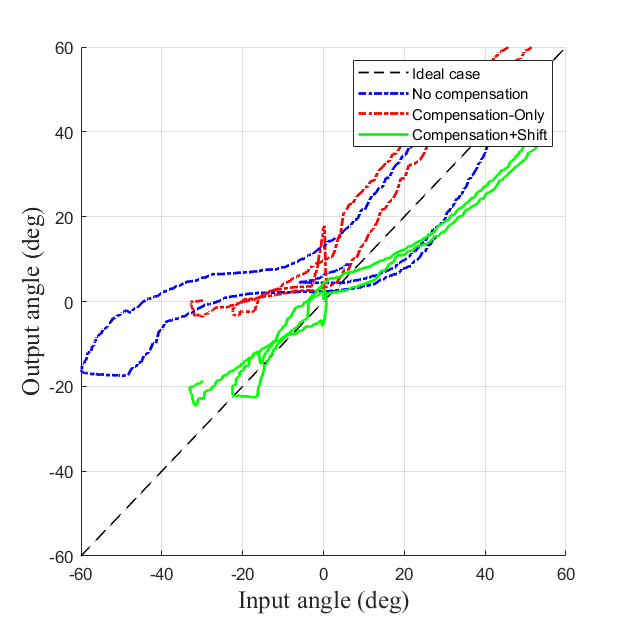}}

		\caption{ This demonstrates 2~DoF results for one catheter: The black dot is the ground truth. The blue dot is {\it No Compensation}. The red dot is {\it Compensation-Only}. The first column is for $\phi_1$. The second column is for $\phi_2$. Both knobs manipulated simultaneously. The first row shows time versus output angle, $\phi_1$ and $\phi_2$ have different frequency as the results demonstrated. The second row shows the input angle versus the output angle for $\phi_1$ and $\phi_2$ motions.
			\label{fig:result_2dof_total}}
	\end{center}
\end{figure}

Figure\,\ref{fig:result_2dof_total} shows 2-DoF results for non-periodic inputs. We setup two curvatures one for AP bending section ($\alpha_1 = 45^{\circ}$, $\theta_1 = 0\circ$) and another for LR bending section $\alpha_2 = 45^{\circ}$, $\theta_2 = 90^\circ$)
The first row shows time versus output angle for $\phi_1$ and $\phi_2$ having different frequency. The second row shows the input angle versus the output angle for $\phi_1$ and $\phi_2$ motions. We have the quantitative measures in Table\,\ref{table:2d_all}. It shows that PTPE is improved by $17.7\%$ to $45\%$ and RMSE is improved by $37.5\%$ to $44\%$.

%

\section{Discussion}

	
The experimental results show that the proposed method is effective to find a shift of hysteresis, and easy to integrate into the basis hysteresis model. Based on \citet{fuxiang10tendon}, the static tendon-sheath force transmission characteristics in curved shape is determined by the tendon curvature, which might change backlash size of hysteresis model. However, based on our experimental results, it is negligible in dynamic motions. Instead, we more focused on how to update the shift of existing models by motor current without external sensors.


The result of two DoF shows also good improvement, however, it does not show as good as one DoF test. Certainly, there exist coupling effects in mechanical structures, which is not detected as much as we expected from motor current. Most commercial products of TDCM has a complicated mechanical structures inside due to multiple purposes ({\em e.g.,} ultrasound image, grasping tools). We think only motor current-based compensation might be limited for more DoF manipulation due to highly nonlinear coupled mechanical system. However, in this paper, we focused on how to compensate shift of hysteresis regarding to shape changes by using motor current without external sensors.

Our proposed method requires a small motion to detect $\tilde{D}_{pos}$ (or $\tilde{D}_{neg}$) in working environments, which can be integrated with the existing hysteresis models. We can think that enough workspace in the bending section will be required for our method. However, from the mechanical properties of dead zone, it is usually located near neutral of knobs. Basically, the dead zone value is less than $20^{\circ}$. Moreover, we can simply notice the first value from the straight condition of model (off-line), thus we can start to search new dead zone from there because next dead zone might be near the old dead zone state.

Our search method requires to detect one side of dead zone.
Thus, at least one side of dead zone should not be colliding with environmental structures, which requires a certain size of workspace for our method. However, if derivation of motor current $\nabla C$ is too high ({\em i.e.} touching the wall), our method certainly detects it based on $\epsilon_{upper}$ boundary. Then, we can simply start to search another direction to find the other side of dead zone. This direction lead the robot go through the dead zone, and reach to other side of dead zone. Actually, due to $\epsilon_{upper}$, our search method is able to generate a safe motion. Overall, we believe it does not require a large workspace to sweep.

\section{Conclusion}	 

This paper introduced a simple method to identify shift of hysteresis due to the arbitrary shape of the proximal shaft. Our gradient-based shift detection method is using motor current as an input, so we do not need external sensors to calibrate this procedure. This method facilitates the robotic system for TDCM to be calibrated in a short time for various environmental structures.

We analyzed our method with several comparison. It showed that 1) The shift of hysteresis is challenging problem, which  might need to be taken into account for hysteresis modeling in practice. Without consideration of shift in hysteresis compensation, the performance might be possible to be worse. 2) The hysteresis shift found by our proposed method can be integrated with any hysteresis models that handles dead zone. 3) The compensator of hysteresis shift significantly improved performance of the tip controls, which were tested in various settings (varied curvature conditions, periodic/non-periodic inputs, 1-DoF and 2-DoF).

\section*{Disclaimer}
The concepts and information presented in this abstract/paper are based on research results that are not commercially available. Future availability cannot be guaranteed.

{
	\small
	\bibliography{references_icra22}

\begin{thebibliography}{22}
\providecommand{\natexlab}[1]{#1}
\providecommand{\url}[1]{#1}
\csname url@samestyle\endcsname
\providecommand{\newblock}{\relax}
\providecommand{\bibinfo}[2]{#2}
\providecommand{\BIBentrySTDinterwordspacing}{\spaceskip=0pt\relax}
\providecommand{\BIBentryALTinterwordstretchfactor}{4}
\providecommand{\BIBentryALTinterwordspacing}{\spaceskip=\fontdimen2\font plus
\BIBentryALTinterwordstretchfactor\fontdimen3\font minus
  \fontdimen4\font\relax}
\providecommand{\BIBforeignlanguage}[2]{{%
\expandafter\ifx\csname l@#1\endcsname\relax
\typeout{** WARNING: IEEEtranN.bst: No hyphenation pattern has been}%
\typeout{** loaded for the language `#1'. Using the pattern for}%
\typeout{** the default language instead.}%
\else
\language=\csname l@#1\endcsname
\fi
#2}}
\providecommand{\BIBdecl}{\relax}
\BIBdecl

\bibitem[Daoud et~al.(1999)Daoud, Kalbfleisch, and Hummel]{daoud1999ep}
E.~Daoud, S.~Kalbfleisch, and J.~Hummel, ``Intracardiac echocardiography to
  guide transseptal left heart catheterization for radiofrequency catheter
  ablation,'' \emph{Journal of Cardiovascular Electrophysiology}, vol.~10,
  no.~3, pp. 358--363, 1999.

\bibitem[Khoshnam and Patel(2017)]{khoshnam2017robotics}
M.~Khoshnam and R.~V. Patel, ``Robotics-assisted control of steerable ablation
  catheters based on the analysis of tendon-sheath transmission mechanisms,''
  \emph{IEEE/ASME Transactions on Mechatronics}, vol.~22, no.~3, pp.
  1473--1484, 2017.

\bibitem[Bai et~al.(2012)Bai, Di~Biase, Valderrabano, Lorgat, Mlcochova, Tilz,
  Meyerfeldt, Hranitzky, Wazni, Kanagaratnam, et~al.]{bai2012worldwide}
R.~Bai, L.~Di~Biase, M.~Valderrabano, F.~Lorgat, H.~Mlcochova, R.~Tilz,
  U.~Meyerfeldt, P.~M. Hranitzky, O.~Wazni, P.~Kanagaratnam \emph{et~al.},
  ``Worldwide experience with the robotic navigation system in catheter
  ablation of atrial fibrillation: methodology, efficacy and safety,''
  \emph{Journal of cardiovascular electrophysiology}, vol.~23, no.~8, pp.
  820--826, 2012.

\bibitem[{Ott} et~al.(2011){Ott}, {Nageotte}, {Zanne}, and {de
  Mathelin}]{ott11endoscopy}
L.~{Ott}, F.~{Nageotte}, P.~{Zanne}, and M.~{de Mathelin}, ``Robotic assistance
  to flexible endoscopy by physiological-motion tracking,'' \emph{IEEE
  Transactions on Robotics}, vol.~27, no.~2, pp. 346--359, 2011.

\bibitem[Le et~al.(2016)Le, Do, and Phee]{le16survey}
H.~M. Le, T.~N. Do, and S.~J. Phee, ``A survey on actuators-driven surgical
  robots,'' \emph{Sensors and Actuators A: Physical}, vol. 247, pp. 323 -- 354,
  2016.

\bibitem[{Dario} and {Mosse}(2003)]{dario03review}
P.~{Dario} and C.~A. {Mosse}, ``Review of locomotion techniques for robotic
  colonoscopy,'' in \emph{2003 IEEE International Conference on Robotics and
  Automation (Cat. No.03CH37422)}, vol.~1, 2003, pp. 1086--1091 vol.1.

\bibitem[Chen et~al.(2006)Chen, Redarce, and Redarce]{chen06kinematics}
G.~Chen, M.~T. Redarce, and T.~Redarce, ``{Development and kinematic analysis
  of a silicone-rubber bending tip for colonoscopy},'' in \emph{Proceedings of
  the International Conference on Intelligent Robots and Systems}, Beijing,
  China, Oct. 2006.

\bibitem[{Phee} et~al.(1997){Phee}, {Ng}, {Chen}, {Seow-Choen}, and
  {Davies}]{phee97locomotion}
S.~J. {Phee}, W.~S. {Ng}, I.~M. {Chen}, F.~{Seow-Choen}, and B.~L. {Davies},
  ``Locomotion and steering aspects in automation of colonoscopy. i. a
  literature review,'' \emph{IEEE Engineering in Medicine and Biology
  Magazine}, vol.~16, no.~6, pp. 85--96, 1997.

\bibitem[Li et~al.(2021)Li, Collins, Kim, Chinnadurai, Mansi, and
  Lin]{zhongyu21ice}
Z.~Li, J.~Collins, Y.-H. Kim, P.~Chinnadurai, T.~Mansi, and C.~H. Lin,
  ``Zero-fluoroscopy transseptal puncture guided by intelligent intracardiac
  echocardiography robotics,'' \emph{Journal of the American College of
  Cardiology}, vol.~77, no. 18\_Supplement\_1, pp. 970--970, 2021.

\bibitem[Kim et~al.(2020)Kim, Collins, Li, Chinnadurai, Kapoor, Lin, and
  Mansi]{kim2020automatic}
Y.-H. Kim, J.~Collins, Z.~Li, P.~Chinnadurai, A.~Kapoor, C.~H. Lin, and
  T.~Mansi, ``Towards automatic manipulation of intra-cardiac echocardiography
  catheter,'' 2020.

\bibitem[Xu and Simaan(2008)]{xu08continuum}
K.~Xu and N.~Simaan, ``An investigation of the intrinsic force sensing
  capabilities of continuum robots,'' \emph{IEEE Transactions on Robotics},
  vol.~24, no.~3, pp. 576--587, 2008.

\bibitem[{Camarillo} et~al.(2008){Camarillo}, {Milne}, {Carlson}, {Zinn}, and
  {Salisbury}]{camarillo08tendon}
D.~B. {Camarillo}, C.~F. {Milne}, C.~R. {Carlson}, M.~R. {Zinn}, and J.~K.
  {Salisbury}, ``Mechanics modeling of tendon-driven continuum manipulators,''
  \emph{IEEE Transactions on Robotics}, vol.~24, no.~6, pp. 1262--1273, 2008.

\bibitem[Rao et~al.(2021)Rao, Peyron, Lilge, and Burgner-Kahrs]{rao21tendon}
P.~Rao, Q.~Peyron, S.~Lilge, and J.~Burgner-Kahrs, ``How to model tendon-driven
  continuum robots and benchmark modelling performance,'' \emph{Frontiers in
  Robotics and AI}, vol.~7, p. 223, 2021.

\bibitem[Robert J. Webster~III(2000)]{webster10ccc}
B.~A.~J. Robert J. Webster~III, ``Design and kinematic modeling of constant
  curvature continuum robots: A review,'' \emph{International Journal of
  Robotics Research}, vol.~29, no.~13, pp. 1661--1683, 2000.

\bibitem[Do et~al.(2014)Do, Tjahjowidodo, Lau, Yamamoto, and
  Phee]{do14hysteresis}
T.~Do, T.~Tjahjowidodo, M.~Lau, T.~Yamamoto, and S.~Phee, ``Hysteresis modeling
  and position control of tendon-sheath mechanism in flexible endoscopic
  systems,'' \emph{Mechatronics}, vol.~24, no.~1, pp. 12 -- 22, 2014.

\bibitem[Xu et~al.(2017)Xu, Poon, Yam, and Chiu]{xu17tendon}
W.~Xu, C.~C.~Y. Poon, Y.~Yam, and P.~W.~Y. Chiu, ``Motion compensated
  controller for a tendon-sheath-driven flexible endoscopic robot,'' \emph{The
  International Journal of Medical Robotics and Computer Assisted Surgery},
  vol.~13, no.~1, p. e1747, 2017.

\bibitem[{Wang} et~al.(2020){Wang}, {Bie}, {Han}, and {Fang}]{wang20hysteresis}
X.~{Wang}, D.~{Bie}, J.~{Han}, and Y.~{Fang}, ``Active modeling and
  compensation for the hysteresis of a robotic flexible ureteroscopy,''
  \emph{IEEE Access}, vol.~8, pp. 100\,620--100\,630, 2020.

\bibitem[Kato et~al.(2014)Kato, Okumura, Kose, Takagi, and Hata]{kato14tendon}
T.~Kato, I.~Okumura, H.~Kose, K.~Takagi, and N.~Hata, ``{Extended Kinematic
  Mapping of Tendon-Driven Continuum Robot for Neuroendoscopy},'' in
  \emph{Proceedings of the International Conference on Intelligent Robots and
  Systems}, Chicago, USA, Sep. 2014.

\bibitem[Zglimbea et~al.(2009)Zglimbea, Finca, Greaban, and
  Constantin]{radu09identification}
R.~Zglimbea, V.~Finca, E.~Greaban, and M.~Constantin, ``Identification of
  systems with friction via distributions using the modified friction lugre
  model,'' in \emph{Proceedings of the 13th WSEAS International Conference on
  Systems}, ser. ICS'09, 2009, p. 579–584.

\bibitem[Hassani and Tjahjowidodo(2013)]{hassani13piezo}
V.~Hassani and T.~Tjahjowidodo, ``Structural response investigation of a
  triangular-based piezoelectric drive mechanism to hysteresis effect of the
  piezoelectric actuator,'' \emph{Mechanical Systems and Signal Processing},
  vol.~36, no.~1, pp. 210 -- 223, 2013.

\bibitem[Lee et~al.(2021)Lee, Kim, Collins, Kapoor, Kwon, and
  Mansi]{lee21hysteresis}
D.-H. Lee, Y.-H. Kim, J.~Collins, A.~Kapoor, D.-S. Kwon, and T.~Mansi,
  ``Non-linear hysteresis compensation of a tendon-sheath-driven robotic
  manipulator using motor current,'' \emph{IEEE Robotics and Automation
  Letters}, vol.~6, no.~2, pp. 1224--1231, 2021.

\bibitem[Fuxiang and Xingsong(2010)]{fuxiang10tendon}
T.~Fuxiang and W.~Xingsong, ``The design of a tendon-sheath-driven robot,''
  \emph{International Journal of Intelligent Systems Technologies and
  Applications}, vol.~8, p. 215, 01 2010.

\end{thebibliography}
}

\end{document}